\documentclass{llncs}
\usepackage[english]{babel}
\usepackage{latexsym,amsmath,amssymb}
\usepackage{xcolor}
\usepackage{graphicx,epsfig,graphics,url,multirow}
\usepackage{algorithm2e}
\usepackage{verbatim,ifthen,array}
\usepackage{multirow,setspace,xspace}
\usepackage{subcaption}
\usepackage{textcomp}
\usepackage{stmaryrd}

\usepackage{booktabs} 
\usepackage{multirow} 
\usepackage{cite}     

\definecolor{citeblue}{rgb}{0.2,0.2,0.7}
\definecolor{urlcolor}{rgb}{0.15,0.15,0.8}
\definecolor{refcolor}{rgb}{0.2,0.2,0.7}
\definecolor{midgreen}{RGB}{0,150,0}
\definecolor{darkgreen}{RGB}{0,128,0}
\definecolor{darkblue}{RGB}{0,0,128}
\definecolor{darkred}{RGB}{192,0,0}
\usepackage[pdftex%
,colorlinks=true%
,bookmarks=true%
,linkcolor=citeblue%
,citecolor=citeblue%
,urlcolor=urlcolor%
,plainpages=false]{hyperref}
\addto\extrasenglish{%

}

\graphicspath{{plots/}}
\usepackage{pgf}
\usepackage{tikz}

\newcommand{\todoF}[2]{}

\newcommand{\var}{\tn{var}} 

\newcommand{\fml}[1]{{\mathcal{#1}}}

\newcommand{\tn}[1]{\textnormal{#1}}

\definecolor{gray}{rgb}{.4,.4,.4}
\definecolor{midgrey}{rgb}{0.5,0.5,0.5}
\definecolor{darkred}{rgb}{0.7,0.1,0.1}
\definecolor{darkblue}{rgb}{0.1,0.1,0.5}
\definecolor{defseagreen}{cmyk}{0.69,0,0.50,0}

\newcounter{Comment}[Comment]
\DeclareMathOperator*{\nentails}{\nvDash}
\DeclareMathOperator*{\entails}{\vDash}

\newcommand{\ddp}{\tn{D}^{\tn{P}}}

\newcommand{\stwop}{\Sigma_2^\tn{P}}

\newcommand{\papdef}{$P=(V,H,M,T,c)$\xspace}
\newcommand{\papdefn}[1]{$P_{#1}=(V,H,M,T,c)$\xspace}

\newcommand{\hyperp}{Hyper$^\star$\,}

\begin{document}

\title{Propositional Abduction with\\ Implicit Hitting Sets}

\author{Alexey Ignatiev \and Antonio Morgado \and
  Joao Marques-Silva\institute{LaSIGE, Faculty of Science,
    University of Lisbon, Portugal, email:
    \{aignatiev,ajmorgado,jpms\}@ciencias.ulisboa.pt}}

\maketitle

%
%
%

\begin{abstract}
  Logic-based abduction finds important applications in artificial 
  intelligence and related areas. One application example is in
  finding explanations for observed phenomena.
  Propositional abduction is a restriction of abduction to the
  propositional domain, and complexity-wise is in the second level of
  the polynomial hierarchy.
  Recent work has shown that exploiting implicit hitting sets and
  propositional satisfiability (SAT) solvers provides an efficient
  approach for propositional abduction.
  This paper investigates this earlier work and proposes a number of 
  algorithmic improvements. These improvements are shown to yield
  exponential reductions in the number of SAT solver calls. 
  More importantly, the experimental results show significant
  performance improvements compared to the the best approaches for
  propositional abduction.
\end{abstract}


%
%
%
%

\section{Introduction} \label{sec:intro}

Logic-based abduction finds relevant applications in artificial
intelligence and related
areas~\cite{bylander-aij91,poole-aij93,marquis-tsmc93,santos-aij94,eiter-jacm95,inoue-cl02,inoue-tplp03,eiter-ds03,satoh-ksem06,zanuttini-sjc06,inoue-ds07,zanuttini-aij08,inoue-ijcai09,woltran-ecai10,inoue-ml13,woltran-jlc15,jarvisalo-kr16}.
Given a background theory and a set of manifestations and a set of
hypotheses, abduction seeks to identify a cost-minimum set of
hypotheses which explain the manifestation and are consistent given
the background theory.
Propositional abduction is hard for the second level of the polynomial
hierarchy, but finds a growing number of
applications~\cite{jarvisalo-kr16,satoh-ksem06}.
Noticeable examples of where propositional abduction algorithms can be applied
include abductive
inference~\cite{reggia-tbe85,dillig-pldi12,dillig-cav13,dillig-oopsla13},
logic programming~\cite{kakas-jlc92,kakas-ilp97,lin-aij02},
knowledge bases updates~\cite{kakas-vldb90,inoue-tplp03},
security protocols verification~\cite{gavanelli-esaw05},
and constraint optimization~\cite{gavanelli-ecai08,gavanelli-iclp09},
among many others.
%

Given the complexity class of propositional abduction, it is
conceptually simple to solve the problem with a linear (or
logarithmic) number of calls to a $\stwop$-oracle, e.g.\ a oracle
for quantified Boolean formulae (QBF) with one quantifier
alternation. Unfortunately, in practice QBF solvers are not as
efficient as SAT solvers, and do not scale as well. As a result,
recent work~\cite{jarvisalo-kr16} on solving propositional abduction
focused on using calls to SAT oracles instead of QBF oracles,
following a trend also observed for solving
QBF~\cite{jms-sat11,jkmsc-sat12,jms-ijcai15,jkmsc-aij16}.
This recent work on solving propositional abduction is motivated by
the practical success of implicit hitting set algorithms in a number
of different
settings~\cite{karp-cmp10,karp-soda11,jms-sat11,bacchus-cp11,stern-aaai12,jkmsc-sat12,karp-or13,liffiton-cpaior13,pms-aaai13,cimatti-sat13,ijms-sat13,jms-ijcai15,amms-sat15,lpmms-cj16,jkmsc-aij16,ijms-cj16,jarvisalo-kr16}.




%
The contributions of this paper can be summarized as follows.
The paper revisits QBF models for solving propositional abduction and
proposes a Quantified MaxSAT (QMaxSAT)~\cite{ijms-sat13,ijms-cj16}
model for propositional abduction. Moreover, the paper notes that the
MaxHS~\cite{bacchus-cp11} approach for MaxSAT can be readily applied
to QMaxSAT by replacing the oracle used.
The paper then investigates the application of implicit hitting sets
to solving propositional abduction~\cite{jarvisalo-kr16} and
identifies a number of algorithmic improvements. This leads to a new
algorithm, Hyper, for solving propositional abduction. This new
algorithm is shown to significantly outperform the current state of
the art, solving a large number of instances that could not be solved
with existing solutions. More importantly, the paper shows that the
algorithmic improvements proposed can save an exponential number of
iterations when compared with the current state of the
art~\cite{jarvisalo-kr16}.

%
The paper is organized as follows.
\autoref{sec:prelim} introduces the definitions used throughout the
paper, and also overviews related work.
\autoref{sec:abd} revisits a QBF model for abduction, which is then
used for developing a number of alternative approaches for solving
propositional abduction. Among these, the paper proposes improvements
to recent work~\cite{jarvisalo-kr16}, which are shown to yield
exponential reductions on the number of SAT oracle calls.
\autoref{sec:res} analyzes the experimental results, running existing
and the proposed algorithms on existing problem
instances~\cite{jarvisalo-kr16}. \autoref{sec:res} also provides
experimental evidence that the proposed algorithms for propositional
abduction can save an exponential number of oracle calls in different
settings.
\autoref{sec:conc} concludes the paper, and identifies possible
research directions.


%
%
%

\section{Preliminaries} \label{sec:prelim}

This section introduces the notation and definitions used throughout
the paper.

\subsection{Satisfiability}

%
%
%
Standard propositional logic definitions apply
(e.g.\ \cite{sat-handbook09}).
CNF formulas are defined over a set of propositional variables.
A CNF formula (or theory) $T$ is a propositional formula represented as a
conjunction of clauses, also interpreted as a set of clauses.
A clause is a disjunction of literals, also interpreted as a set of
literals.
A literal is a variable or its complement.
The set of variables of a theory $T$ is denoted $X\triangleq\var(T)$.
The dependency of $T$ on $X$ can be made explicit by writing $T(X)$.
Where convenient, a formula can be rewritten with a fresh set of
variables, e.g.\ we can replace $X$ by $Y$, writing $T(Y)$.
Conflict-driven clause learning (CDCL) SAT solvers are summarized
in~\cite{sat-handbook09}.
Throughout the paper, SAT solvers are viewed as oracles. Given a CNF formula
$F$, a SAT oracle decides whether $F$ is satisfiable, and returns a satisfying
assignment $\mu$ if $F$ is satisfiable. A SAT oracle can also return a subset
of the clauses (i.e.\ an unsatisfiable core $U\subseteq F$) if $F$ is
unsatisfiable.

CNF formulas are often used to model overconstrained problems.
In general, clauses in a CNF formula are characterized as hard,
meaning that these must be satisfied, or soft, meaning that these are
to be satisfied, if at all possible. A weight can be associated with
each soft clause, and the goal of maximum satisfiability (MaxSAT) is
to find an assignment to the propositional variables such that the
hard clauses are satisfied, and the sum of the satisfied soft clauses
is maximized. Branch-and-bound algorithms for MaxSAT are overviewed
in~\cite{sat-handbook09}.
Recent work on MaxSAT investigated core-guided
algorithms~\cite{ansotegui-aij13,mhlpms-cj13} and also the use of
implicit hitting sets~\cite{bacchus-cp11}.

In the analysis of unsatisfiable CNF formulas, a number of definitions
are used.
Given an unsatisfiable CNF formula $F$, a minimal unsatisfiable subset
(MUS) $M\subseteq F$ is both unsatisfiable and irreducible.
Given an unsatisfiable CNF formula $F$, a minimal correction subset
(MCS) $C\subseteq F$ is both irreducible and its complement is
satisfiable.
Given an unsatisfiable CNF formula $F$, a maximal satisfiable subset
(MSS) $S$ is the complement of some MCS of $F$.
A largest MSS is a solution to the MaxSAT problem.

Additionally, it is well-known~\cite{reiter-aij87,liffiton-jar08} that MUSes
and MCSes are connected by a \emph{hitting set duality}.
Given a collection $\Gamma$ of sets from a universe $\mathbb{U}$, a hitting
set $h$ for $\Gamma$ is a set such that $\forall S \in \Gamma, h
\cap S \ne \emptyset.$
A hitting set $h$ is \emph{minimal} if none of its subset is a hitting set.
It is straightforward to extend the previous definitions to the case
where there are hard clauses.

\emph{Quantified Boolean formulas} (QBFs) are an extension of propositional
logic with \emph{existential} and \emph{universal} quantifiers ($\forall$,
$\exists$)~\cite{sat-handbook09}.
A QBF can be in \emph{prenex closed} form $Q_1{x_1}{\dots}Q_n{x_n}.\ \varphi$,
where $Q_i\in\{\forall, \exists\}$, $x_i$ are distinct Boolean variables, and
$\varphi$ is a Boolean formula over the variables $x_i$ and the constants $0$
(false), $1$ (true). The sequence of quantifiers in a QBF is called the {\em
prefix} and the Boolean formula the {\em matrix}.
The semantics of QBF is defined recursively. A QBF
$\exists{x_1}Q_2{x_2}{\dots}Q_n{x_n}.\,\varphi$ is true iff
$Q_2{x_2}{\dots}Q_n{x_n}.\ \varphi|_{x_1=1}$  or
$Q_2{x_2}{\dots}Q_n{x_n}.\ \varphi|_{x_1=0}$  is true.
A QBF $\forall{x_1}Q_2{x_2}{\dots}Q_n{x_n}.\,\varphi$ is true iff
both $Q_2{x_2}{\dots}Q_n{x_n}.\ \varphi|_{x_1=1}$ and
$Q_2{x_2}{\dots}Q_n{x_n}.\ \varphi|_{x_1=0}$  are true.
To decide whether a given QBF is true or not, is known to be
{\sf PSPACE}-complete~\cite{sat-handbook09}.

\subsection{Propositional Abduction}

A propositional abduction problem (PAP) is a 5-tuple \papdef.
$V$ is a finite set of variables. $H$, $M$ and $T$ are CNF
formula representing, respectively, the set of hypotheses, the set of
manifestations, and the background theory.
$c$ is a cost function associating a cost with each clause of $H$,
$c\::\:H\to\mathbb{R}^{+}$.

Given a background theory $T$, a set $S\subseteq H$ of hypotheses is
an explanation (for the manifestations) if: (i) $S$ entails the
manifestations $M$ (given $T$); and (ii) $S$ is consistent (given $T$).
%
The propositional abduction problem consists in computing a minimum size
explanation for the manifestations subject to the background theory.

\begin{definition}[Explanations for $P$~\cite{jarvisalo-kr16}]
Let \papdef be a PAP. The set of explanations of $P$ is given by the
set $\tn{Expl}(P) = \{S\subseteq H\:|\: T\land S\nentails\bot ,
T\land S\entails M\}$.
The minimum-cost solutions of $P$ are given by $\tn{Expl}_c(P) =
\tn{argmin}_{E\in\tn{Expl}(P)}(c(E))$.
\end{definition}

The complexity of logic-based abduction has been investigated in a
number of works~\cite{bylander-aij91,eiter-jacm95}, and is surveyed
in~\cite{jarvisalo-kr16}.
Checking whether $S\subseteq H$ is an explanation for a PAP is
$\ddp$-complete. Deciding the existence of some explanation is
$\stwop$-complete. Finding a minimum-size explanation can be achieved
with a linear number of calls to a $\stwop$ oracle or, if the costs
are polynomially bounded, with a logarithmic number of calls to a
$\stwop$ oracle.

\begin{example}[Example abduction instance.] \label{ex:abd01}
Consider the propositional abduction problem instance \papdef with
the set of variables $V$,
the set of hypotheses $H$,
the manifestations $M$, and
the background theory $T$ given by,
\begin{equation} \label{eq:ex01}
  \begin{array}{ll}
    V = & \{x_1, x_2, x_3, x_4 \} \\
    H = & \{(x_1), (x_2), (x_3)\} \\
    M = & \{(x_4)\} \\
    T = & \{(\neg x_1\vee x_4),(\neg x_2\vee\neg x_3\vee x_4)\} \\
  \end{array}
\end{equation}
The (propositional) abduction problem for this example is to find a
minimum cost subset $S$ of $H$, such that (i) $S$ is consistent with
$T$ (i.e.\ $T\land S\nentails\bot$); and (ii) $S$ and $T$ entail $M$
(i.e.\ $T\land S\entails M$).
For this instance of propositional abduction, the  minimum cost
explanation is then $S = \{(x_1)\}$.
\end{example}


%
%
%

\subsection{Related Work} \label{ssec:relw}

This paper builds on recent work on algorithms for propositional
abduction~\cite{jarvisalo-kr16}, which builds on earlier work on
solving maximum satisfiability with implicit hitting set
algorithms~\cite{bacchus-cp11}.
This work is also tightly related with the body of work on handling
hitting sets
implicitly~\cite{karp-cmp10,karp-soda11,bacchus-cp11,karp-or13,cimatti-sat13,jarvisalo-kr16},
which is also tightly related with implicit hitting set
dualization~\cite{stern-aaai12,liffiton-cpaior13,pms-aaai13,amms-sat15,lpmms-cj16},
but also with abstraction refinement in QBF solving and
optimization~\cite{jms-sat11,jkmsc-sat12,ijms-sat13,jms-ijcai15,jkmsc-aij16,ijms-cj16}.

The use of implicit hitting sets for solving MaxSAT is embodied by
MaxHS (e.g.\ see~\cite{bacchus-cp11}), which is summarized in~\autoref{alg:mxhs}.
\begin{algorithm}[t]
%
%
%
%
%
%
%

\DontPrintSemicolon
\SetAlgoNoLine
\LinesNumbered
\SetFillComment
\SetKw{KwNot}{not\xspace}
\SetKw{KwAnd}{and\xspace}
\SetKw{KwOr}{or\xspace}
\SetKw{KwBreak}{break\xspace}
\SetKwData{false}{{\small false}}
\SetKwData{true}{{\small true}}
\SetKwData{st}{{\slshape st}}
\SetKwData{cores}{$\mathcal{C}$}
\SetKwFunction{SAT}{SAT}
\SetKwFunction{CNF}{CNF}
\SetKwFunction{minhs}{MinimumHS}
\SetKwFunction{getmus}{ComputeMUS}
\SetKwFunction{relaxcls}{RelaxCls}
\SetKwFunction{softcls}{SoftCls}
\SetKwFunction{falsecls}{PickFalseCls}
\SetKwFunction{relaxvars}{GetRelaxationVars}
\SetKwFunction{ip}{IP}
\SetKwFunction{cost}{Cost}
\SetKwBlock{Let}{let}{end}
\SetKwBlock{FBlock}{}{end}

\KwIn{$F$ WCNF formula}
\KwOut{$(\mu, \cost(\mu))$ MaxSAT assignment and cost}
\Begin{
  \SetAlgoVlined
  $K\gets \emptyset$ \;
  \While{\true}{
    $h \gets \minhs(K)$ \;
    $(\st,\mu)\gets\SAT(F\setminus h)$ \;
    \tcp{If $\st$, then $\mu $ is an assignment}
    \tcp{Otherwise, $\mu$ is a core}
    \lIf{$\st$}{\Return $(\mu, \cost(\mu))$}
    $K\gets K \cup \{\mu\}$\;
  }
}
\BlankLine
%

  \caption{The MaxHS algorithm~\cite{bacchus-cp11}} \label{alg:mxhs}
\end{algorithm}
The algorithm computes minimum hitting sets of a set of sets, each of
which represents an unsatisfiable subformula of the target formula.
This essentially exploits Reiter's~\cite{reiter-aij87} well-known
hitting set relationship between MCSes and
MUSes~\cite{lozinskii-jetai03,stuckey-padl05,liffiton-jar08}, where an
MCS is a minimal hitting set of the MUSes and vice-versa. Moreover,
since a minimum hitting set is being computed, we are in search of the
smallest MCS, i.e.\ the MaxSAT solution.
In case there are hard clauses, MaxHS needs to take this into
consideration, including checking the consistency of the hard
clauses.
Besides MaxHS, recent work on MaxSAT solving is based on iterative
unsatisfiable core identification~\cite{ansotegui-aij13,mhlpms-cj13}.

Recent work on propositional abduction builds on MaxHS and proposes a
novel algorithm, AbHS/AbHS+. AbHS mimics MaxHS in that the algorithm
iteratively computes minimum cost hitting sets, which identify a
subset $S\subseteq H$. This set $S$ is then used for checking
whether it represents an explanation of the propositional abduction
problem. Since it is a minimum hitting set then, if the conditions
hold, it is a minimum-cost explanation.
%
AbHS is summarized in~\autoref{alg:abhs}.
\begin{algorithm}[t]
%
%
%
%
%
%
%

\DontPrintSemicolon
\SetAlgoNoLine
\LinesNumbered
\SetFillComment
\SetKw{KwNot}{not\xspace}
\SetKw{KwAnd}{and\xspace}
\SetKw{KwOr}{or\xspace}
\SetKw{KwBreak}{break\xspace}
\SetKwData{false}{{\small false}}
\SetKwData{true}{{\small true}}
\SetKwData{st}{{\slshape st}}
\SetKwData{cores}{$\mathcal{C}$}
\SetKwFunction{SAT}{SAT}
\SetKwFunction{CNF}{CNF}
\SetKwFunction{minhs}{MinimumHS}
\SetKwFunction{getmus}{ComputeMUS}
\SetKwFunction{relaxcls}{RelaxCls}
\SetKwFunction{softcls}{SoftCls}
\SetKwFunction{falsecls}{PickFalseCls}
\SetKwFunction{relaxvars}{GetRelaxationVars}
\SetKwFunction{ip}{IP}
\SetKwFunction{cost}{Cost}
\SetKwBlock{Let}{let}{end}
\SetKwBlock{FBlock}{}{end}

\KwIn{PAP \papdef}
\KwOut{Minimum cost explanation $S$}
\Begin{
  \SetAlgoVlined
  $K\gets \emptyset$ \;
  $S\gets \emptyset$ \;
  $(\st,\mu)\gets\SAT(T\land H\land(\neg M))$ \;
  \lIf{$\st$}{\Return $\emptyset$}
  \While{$S \neq H$}{
    $(\st,\mu)\gets\SAT(T\land S\land(\neg M))$ \;
    \lIf{$\st$}{$K\gets K \cup \{\{h\in H ~|~ \mu(h) = 0\}\}$}
    \Else{
      $(\st,\mu)\gets\SAT(T\land S)$ \;
      \lIf{\KwNot$\st$}{$K\gets K \cup \{(H\setminus S)\}$}
      \lElse{\Return S} 
    }
    $S\gets\minhs(K)$ \label{ln:ilp}\;
  }
  \Return $\emptyset$ \;
}
\BlankLine
%

  \caption{The AbHS/AbHS+ abduction algorithm~\cite{jarvisalo-kr16}} \label{alg:abhs}
\end{algorithm}
As in MaxHS, the algorithm iteratively computes minimum hitting sets
using an ILP solver (\autoref{ln:ilp}). The outcome is a subset $S$ of
$H$.
The abduction conditions are checked with two distinct SAT oracle
calls. One oracle call checks whether $T\land S\land(\neg M)$ is
inconsistent, i.e.\ whether $T\land S\entails M$. If the formula is
satisfiable, then the set of sets to hit ($K$) is updated with another
set, of the clauses in $H$ falsified by the computed satisfying
assignment. If $T\land S\land(\neg M)$ is inconsistent, then a second
oracle call checks whether $T\land S$ is consistent. If it is, then a
minimum-cost explanation has been identified. Otherwise, AbHS creates
a hitting set by requiring that some non-selected clause of $H$ be
selected in subsequently computed minimum hitting sets.
For the AbHS+ variant~\cite{jarvisalo-kr16}, the set added can be
viewed as the complements of the literals selecting each clause in
$S$, i.e.\ at least some clause in $S$ must not be picked.


There is a vast body of work on exploiting implicit hitting sets.
The concept of exploiting implicit hitting sets is intended to mean
that, instead of starting from an explicit representation of the
complete set of hitting sets, hitting sets are computed on demand, as
deemed necessary by the problem being solved.
An earlier example of exploiting implicit hitting sets is the work of
Bailey and Stuckey~\cite{stuckey-padl05}, in the concrete application
to hitting set dualization. The concept was re-introduced more
recently~\cite{karp-cmp10,karp-soda11,bacchus-cp11}, and then applied
in a number of different settings.
Among this vast body of work, as will become clear throughout the
paper, our work can be related with abstraction refinement ideas used
in recent expansion-based QBF solvers, namely
RAReQS~\cite{jms-sat11,jkmsc-sat12,ijms-cj16} and quantified
optimization extensions~\cite{ijms-sat13,jkmsc-aij16}, but now in the
context of handling implicit hitting sets.


%
%
%

\section{Algorithms for Propositional Abduction} \label{sec:abd}

This section overviews different algorithms for solving propositional
abduction, all of which are based on reducing the problem to QBF.

\subsection{QBF Model for Abduction} \label{ssec:qbfm}

Given a PAP \papdef, the problem of deciding whether some set $S$ is
an explanation can be reduced to QBF.
$S\subseteq H$ is an explanation of $P$ iff:
\begin{equation} \label{eq:qpap}
  \exists_{X} T(X)\land S(X) \land \forall_Y \neg(T(Y)\land
  S(Y)\land\neg M(Y))
\end{equation}
is true. (Observe $X$ and $Y$ denote sets of variables, thus
highlighting that different sets of variables are used.)
\eqref{eq:qpap} can be rewritten as follows:
\begin{equation} \label{eq:qpap2}
  \exists_{X}\phi(X)\land\forall_Y\psi(Y),
\end{equation}
where $\phi=T\land S$ and $\psi=\neg(T\land S\land\neg M)$.

As indicated in~\autoref{sec:prelim}, the goal of propositional
abduction is to find a minimum cost explanation, i.e.\ to pick a
minimum cost set $S\subseteq H$ that is an explanation of $P$.

The problem of finding a minimum cost explanation of $P$ can be
reduced to quantified maximum satisfiability~\cite{ijms-cj16}
(QMaxSAT). Associate a variable $r_i$ with each clause $C_i\in H$,
and create a set $H'$ where each clause $C_i\in H$ is replaced by
$(r_i\lor C_i)$, to enable relaxing the clause. Let $R$ denote
the set of the $r_i$ (relaxation) variables, with $|R| = |H|$.
$H'$ serves to create a modified QBF:
\begin{equation} \label{eq:qmxpap}
  \exists_R\exists_X T(X)\land H'(R,X)\land \forall_Y \neg(T(Y)\land
  H'(R,Y)\land\neg M(Y))
\end{equation}
As before,~\eqref{eq:qmxpap} can be rewritten as follows:
\begin{equation} \label{eq:qmxpap2}
  \exists_R\exists_X\phi(X,R)\land\forall_Y\psi(Y,R)
\end{equation}
The above QBF can be transformed into prenex normal formal, and
represents the hard part of the QMaxSAT problem.
Moreover, the fact that the goal is to compute a minimum cost
explanation of $P$ is modeled by adding a soft clause $(\neg r_i)$,
with cost $c(C_i)$, for each $r_i\in R$.
Each soft clause denotes a preference not to include the associated
clause in $H$ in the computed explanation.

\begin{example}
  With respect to the PAP from example~\autoref{ex:abd01}, the
  QBF associated with the hard part of the QMaxSAT problem is:
\begin{equation}
  \begin{array}{l}
    \exists_{r_1,r_2,r_3}\exists_{x_1, x_2, x_3, x_4} \\
    \quad (\neg x_1\vee x_4)\land(\neg x_2\vee\neg x_3\vee x_4)\land \\
    \quad (r_1\lor x_1)\land(r_2\lor x_2)\land(r_3\lor x_3) \\
    \forall_{y_1, y_2, y_3, y_4} \\
    \quad \neg [ 
      (\neg y_1\vee y_4)\land(\neg y_2\vee\neg y_3\vee y_4)\land \\
    \quad \quad (r_1\lor y_1)\land(r_2\lor y_2)\land(r_3\lor
      y_3)\land(\neg y_4) ]\\ 
  \end{array}
\end{equation}
with the soft clauses being $\{(\neg r_1),(\neg r_2),(\neg r_3)\}$.
\end{example}

The QMaxSAT formulation can be used to develop a number of alternative
approaches for solving PAP. These approaches are detailed in the next
sections.

Observe that, if the propositional abduction problem is not trivial to
solve, then the QBF~\eqref{eq:qpap} is false for $S=\emptyset$ and
$S=H$.
%
%
%

\subsection{Abduction with QMaxSAT}

Similarly to MaxSAT, a number of algorithms can be envisioned for
solving QMaxSAT. These are analyzed in the subsections below, taking
into account the specific structure of the reduction of propositional
abduction to QMaxSAT.

\subsubsection{Iterative QBF Solving}

A standard approach for solving MaxSAT is iterative SAT
solving~\cite{mhlpms-cj13}. Similarly, we can use iterative QBF
solving for QMaxSAT.
At each step, and given cost $k$, the following pseudo-Boolean
constraint is used:
\begin{equation}
\tn{PB}(R,k)\triangleq\left(\sum_{C_i\in H} c(C_i)r_i\le k\right)
\end{equation}
For some positive $k$, the QBF~\eqref{eq:qmxpap2} can be used for
iteratively QBF solving as follows:
%
\begin{equation} \label{eq:qmxpap3}
  \exists_R\tn{PB}(R,k)\land\exists_X\phi(X,R)\land\forall_Y\psi(R,Y)
\end{equation}

Clearly, binary search can be used to ensure a linear (or logarithmic,
depending on whether the costs are bounded) number of $\stwop$ oracle
calls~\cite{eiter-jacm95}.
In practice, most QBF solvers expect clausal representations. Clausification
introduces one additional level of quantification. Typically, each
quantification level makes a QBF formula harder to decide.
%
%
And thus in practice, QBF solvers scale worse than SAT solvers, and so this
approach is unlikely to scale for large propositional abduction
problem instances.

\subsubsection{Core-Guided QBF Solving}

Core-guided algorithms~\cite{ansotegui-aij13,mhlpms-cj13} represent
another approach for solving QMaxSAT.
Many variants of core-guided MaxSAT algorithms have been proposed in
recent years~\cite{ansotegui-aij13,mhlpms-cj13}.

Given the reduction of propositional abduction to QMaxSAT, any
core-guided Max\-SAT algorithm can be used, provided a core-producing
QBF solver is used~\cite{ijms-cj16,egly-sat15}.

Nevertheless, for the abduction problem the use of alternative MaxSAT
solving approaches, based on MaxHS~\cite{bacchus-cp11} is amenable to
efficient optimizations, which solely use SAT
solvers~\cite{jarvisalo-kr16}.

\subsubsection{Exploiting MaxHS}

A recent approach for MaxSAT solving is MaxHS~\cite{bacchus-cp11},
which exploits integer linear programming (ILP) solving, resulting in
simpler SAT oracle calls, at the cost of a possibly exponentially
larger number of oracle calls.
Recall that the MaxHS approach for solving MaxSAT is outlined
in~\autoref{alg:mxhs}.


A straightforward solution for solving QMaxSAT is to replace the SAT oracle by
a QBF oracle in MaxHS. This approach is referred to as QMaxHS, and it was
implemented on top of DepQBF, the known QBF solver which is capable of
reporting unsatisfiable cores if the input QBF is false~\cite{egly-sat15}. It
should be noted (and it is also mentioned in~\autoref{sec:res}) that the
implementation of QMaxHS performs quite bad (it cannot solve any benchmark
instances considered in~\autoref{sec:res}). A possible explanation of this is
that the QBF formulas, which are iteratively solved by the QBF solver, are too
hard even though the original idea of MaxHS-like algorithms is to get (many)
simple calls to the oracle. This suggests that implementing core-guided QMaxSAT
algorithms would not pay off as well since the QBF formulas in core-guided
QMaxSAT are much harder to deal with.
Recent work on propositional abduction~\cite{jarvisalo-kr16} proposed a
MaxHS-like approach, but the QBF oracle call was replaced by two SAT
solver calls, which is expected to outperform QMaxHS.

%

%
%
%

\subsection{Exploiting Implicit Hitting Sets} \label{ssec:ihs}


The use of implicit hitting sets for abduction was proposed in recent
work~\cite{jarvisalo-kr16}. This work can be viewed as extending the
MaxHS algorithm for MaxSAT~\cite{bacchus-cp11}, which is based on
implicit enumeration of hitting sets. In contrast to MaxHS, instead
of one SAT oracle call, AbHS~\cite{jarvisalo-kr16} uses two SAT oracle
calls, one to check entailment of $M$ by $T\land S$ and another to
check the consistency of $T\land S$.
A variant of AbHS, AbHS+, differs on which sets are added to the
hitting set representation.
Whereas the connection of MaxHS and AbHS with implicit hitting sets is
clear, the approach used in AbHS+ can be viewed as adding both only
positive clauses and only negative clauses to hit, and so the
connection with hitting sets is less evident.
An alternative way of explaining AbHS/AbHS+ is to
consider~\eqref{eq:qmxpap2}. The ILP solver is used for computing some
minimum cost hitting set, which represents a set of clauses
$S\subseteq H$. Then, one SAT oracle call checks $\exists_{X}\phi(X)$,
given $S$, and another SAT oracle call checks $\forall_Y\psi(Y)$, also
given $S$. (Observe that this second formula corresponds to checking
unsatisfiability.)
This explanation of how AbHS/AbHS+ works is investigated in greater
detail below.

In the following an alternative approach for propositional
abduction is developed which, similarly to AbHS/AbHS+, is also based
on handling implicit hitting sets, but which is shown to yield
exponential reductions on the number of oracle calls in the worst
case.
The new algorithm, Hyper, shares similarities with MaxHS and also with
AbHS/AbHS+ in that minimum hitting sets are also computed, and an
implicit representation of the hitting sets is maintained. However,
and in contrast with AbHS/AbHS+, Hyper analyzes the structure of the
problem formulation, and develops a number of optimizations that
exploit that formulation.

The propositional abduction problem formulation can be presented in a
slightly modified form, i.e.\ to find a smallest cost set $S\subseteq
H$, consistent with $T$ (i.e.\ a $T$-consistent set $S$), which,
together with $T$, entails $M$.
Consider a $T$-consistent candidate set $S\subseteq H$. If $T\land
S\nentails M$, then the formula $T\land
S\land (\neg M)$ is satisfiable and the satisfying assignment returned by the
SAT oracle is a {\em counterexample} explaining why the selected set
$S$ is such that $T\land S\nentails M$. Moreover, this satisfying
assignment can be used for revealing a (possibly subset-minimal) set
of clauses in $H\setminus S$ which are falsified.
Clearly, one of these falsified clauses must be included (i.e. {\em
  hit}) in any $T$-consistent set $S\subseteq H$ which, together with
$T$, will entail $M$.
Thus, from each $T$-consistent candidate set $S\subseteq H$ which,
together with $T$, does not entail $M$, we can identify a set of
clauses, from which at least one must be picked, in order to pick
another $T$-consistent set $S\subseteq H$, such that eventually
$T\land S\entails M$.

The approach outlined in the paragraph above, although apparently
similar to the description of AbHS/AbHS+, reveals a number of
significant insights. First, checking the consistency of $S$ with $T$
can be carried out {\em concurrently} with the selection of $S$
itself. This is also apparent from the QBF
formulation~\eqref{eq:qpap}, in that all existential quantifiers can
be aggregated and handled simultaneously.
Thus, each minimum hitting set $S$ is computed while guaranteeing that
$T\land S$ holds.
More importantly, after each set $S$ is picked, it is only necessary
to check whether $T\land S\entails M$, and this can be done with a
{\em single} SAT oracle call.

Concretely, the next minimum cost hitting set, given the already
identified sets to hit, is computed guaranteeing that the existential
part of~\eqref{eq:qmxpap} is satisfied:
\begin{equation} \label{eq:papexists}
  \exists_R\exists_X T(X)\land H'(R,X)
\end{equation}
The selected set $S$, identified by the assignment to the $R$
variables, is then used for checking the satisfiability of the second
component of QBF~\eqref{eq:qpap}:
\begin{equation}
  \forall_Y \neg(T(Y)\land S(Y)\land\neg M(Y))
\end{equation}
Observe that this can be decided with a SAT oracle call.

Thus, a careful analysis of the problem formulation enables improving
upon AbHS and AbHS+, specifically by eliminating one SAT oracle call per
iteration. However, as shown in the next section, the new algorithm
can save an exponentially large number of SAT oracle calls when
compared with AbHS/AbHS+.

Besides the aggregation of the existential quantifiers, additional
optimizations can be envisioned.
Propositional abduction seeks a minimum cost set $S\subseteq H$ such
that $T\land S\entails M$. Ideally, one would prefer not to select a
set $S\subseteq H$ such that $T\land S\nentails M$. Observe that
$T\land S\entails M$ implies that $T\land S\land M$ holds, but the
converse is in general not true. Thus,  $M$ can be added
to~\eqref{eq:papexists}, resulting in requiring that $T\land S\land M$
be consistent when selecting the set $S$.
The inclusion of $M$ when picking a minimum hitting set can also
reduce the number of oracle calls exponentially. This is also
investigated in the next section.

The new Hyper algorithm for propositional abduction is shown
as~\autoref{alg:cexr}.
\begin{algorithm}[t]
%
%
%
%
%
%
%

\DontPrintSemicolon
\SetAlgoNoLine
\LinesNumbered
\SetFillComment
\SetKw{KwNot}{not\xspace}
\SetKw{KwAnd}{and\xspace}
\SetKw{KwOr}{or\xspace}
\SetKw{KwBreak}{break\xspace}
\SetKwData{false}{{\small false}}
\SetKwData{true}{{\small true}}
\SetKwData{st}{{\slshape st}}
\SetKwData{cores}{$\mathcal{C}$}
\SetKwFunction{SAT}{SAT}
\SetKwFunction{CNF}{CNF}
\SetKwFunction{minhs}{MinimumHS}
\SetKwFunction{getmus}{ComputeMUS}
\SetKwFunction{relaxcls}{RelaxCls}
\SetKwFunction{softcls}{SoftCls}
\SetKwFunction{falsecls}{PickFalseCls}
\SetKwFunction{relaxvars}{GetRelaxationVars}
\SetKwFunction{ip}{IP}
\SetKwFunction{cost}{Cost}
\SetKwBlock{Let}{let}{end}
\SetKwBlock{FBlock}{}{end}

\KwIn{PAP \papdef}
\KwOut{Minimum cost explanation $S$}
\Begin{
  \SetAlgoVlined
  $(H',R)\gets\relaxcls(H)$ \;
  $B\gets T\land M\land H'$ \;
  $A\gets \emptyset$ \;
  \While{\true}{
    $(\st, h)\gets\minhs(A, B)$ \label{ln:maxmod} \;
    \lIf{\KwNot$\st$}{\Return $\emptyset$}
    $S\gets \{ C_i\in H'\,|\,r_i\in h\}$ \;
    $(\st,\mu)\gets\SAT(T\land S\land(\neg M))$ \;
    \lIf{\KwNot$\st$}{\Return $S$}
    $W\gets\falsecls(H\setminus S, \mu)$ \;
    $Y\gets\relaxvars(W)$ \;
    $A\gets A\cup Y$ \;
  }
}
\BlankLine
%

  \caption{Organization of Hyper} \label{alg:cexr}
\end{algorithm}
Clauses in $H$ are relaxed, to allow each clause $C_i\in H$ to be
picked when the associated relaxation variable $r_i$ is assigned value
1.
The minimum hitting sets are computed for the set of sets $A$, subject
to a background theory $B$, which conjoins $T$, $M$ and the relaxed
clauses of $H$.
In Hyper minimum hitting sets are computed with a MaxSAT
solver~\cite{mims-jsat15}, since the hard part (containing $T$, $M$
and $H'$) plays a significant role in deciding consistency.
If all hitting sets have been (implicitly) tried unsuccessfully, then
the algorithm terminates and returns $\emptyset$. If not, and if
$T\land S\entails M$, then the algorithm terminates and returns the
computed set $S$. Otherwise, a subset of the clauses in $H\setminus
S$, which are falsified by the computed satisfying assignment $\mu$,
is identified, and the associated relaxation variables are used to
create another set to hit, i.e.\ one of those clauses must be included
in any selected set $S$.

\subsubsection{Additional Optimizations} \label{sec:opt}

A few additional optimizations are possible, which can be expected to
have some impact in the performance of Hyper. These are discussed
next.
The first two optimizations are implemented in a variant of Hyper,
Hyper$^\star$. The other optimizations are analyzed to explain why
performance improvements are not expected to be significant.

One optimization is to perform \emph{partial reduction} of the
counterexamples computed in lines 11--12 of \autoref{alg:cexr}. Recall
that counterexamples, i.e.\ sets that need to be hit next time,
comprise clauses of $H\setminus S$ that are falsified by a model $\mu$
of $T\land S\land (\neg{M})$. Thus, the counterexamples can be seen as
correction subsets for the partial CNF formula $T\land H\land
(\neg{M})$, where $T\land(\neg{M})$ is the hard part and $H$ is the
soft part. Observe that instead of computing \emph{any} correction
subset, one may want to reduce it to get a subset-minimal correction
subset (an MCS), i.e.\ to try to minimize the number of falsified
clauses in $H\setminus S$. In \hyperp this is done using the standard
linear search algorithm~\cite{mshjpb-ijcai13}, which iterates through
the falsified clauses and tries to satisfy them. In order not to spend
too much time on doing the reduction, the version of \hyperp presented
here iterates only over $0.2\times
m$ falsified clauses of the initial counterexample starting from the clauses of
the smallest weight, where $m$ is the size of the initial counterexample.

A second optimization is to start by computing a fixed number of
minimum hitting sets. Given that $T\land H\land (\neg M)$ is
inconsistent, one can enumerate MCSes of this formula, which must be
hit, so that one can eventually prove that there exists some set $S$
with $T\land S\entails M$.
In Hyper, 100 MCSes of $T\land H\land (\neg M)$ are computed before
starting the process of generating candidate sets $S$. These MCSes are
computed by size, using MaxSAT-based MCS
enumeration~\cite{liffiton-jar08,mlms-hvc12}.

A third optimization respects the clauses in $M$. Any $C_j\in M$, such
that $T\entails C_j$, can be removed from $M$.
In a preprocessing step, each clause in $M$ is checked for entailment
with respect to $T$. Any entailed clause is removed.
This technique reduces the practical hardness of the formulas checked
for unsatisfiability. Since most of the running time of Hyper is spent
on computing minimum hitting sets, the impact of the technique is
expected to be marginal.

A fourth, and final optimization respects the clauses in $H$. Any
$C_j\in H$ that $T\entails C_j$, can also be removed from $H$, as it
will not be included in any minimum cost hitting set.
It should be noted that the gains of this technique should be also
marginal. Since by construction $T\land S$ is consistent, and the
only computed counterexamples satisfy $T\land S\land(\neg M)$, then
any clause $C_j\in H$ with $T\entails C_j$ will also be satisfied.
Since, the counterexamples only consider falsified clauses, then any
clause $C_j\in H$ entailed by $T$ will {\em never} be included in a
set to be hit.

\subsubsection{Exponential Reductions in Oracle Calls} 
\label{sec:analysis}

This section argues that the new Hyper algorithm for solving
propositional abduction can save an exponentially larger number of
iterations when compared with the AbHS/AbHS+ algorithm proposed
in earlier work~\cite{jarvisalo-kr16}.

Consider a PAP \papdefn{1}, with:
\begin{equation} \label{eq:expex1}
\begin{array}{ll}
  V = & \{ t_1, x_1, y_1, m_1, \ldots, t_n, x_n, y_n, m_n \} \\
  H = & \{ (\neg x_1), (x_1\lor t_1), (\neg y_1), (y_1\lor t_1),\ldots, \\
  & (\neg x_n), (x_n\lor t_n), (\neg y_n), (y_n\lor t_n) \} \\
  M = & \{ (m_1), (m_2), \ldots, (m_n) \} \\
  T = & \{ (\neg t_1\lor\neg t_2\lor\ldots\lor\neg t_n),\\
  & (\neg t_1\lor m_1),\ldots,(\neg t_n\lor m_n) \} \\
\end{array}
\end{equation}
%
and $c(C_i) = 1$ for $C_i\in H$.
Clearly, $P$ has no solution. %
For $M$ to be entailed, $S$ must imply all variables $t_i$ to 1; but
this causes $T\land S$ to become inconsistent.
Moreover, there are exponentially many sets $S$, which are not
consistent with $T$. AbHS+ will have to enumerate all of these sets
$S$ and, for each such set $S$, it will use one additional SAT oracle
call to conclude that $T\land S$ is inconsistent.
Since AbHS+ (or AbHS) selects all falsified clauses when blocking
counterexamples of $T\land S\land(\neg M)$, all subsets of $S$
inconsistent with $T$ will be eventually enumerated.
In contrast, since Hyper ensures consistency between $S$ and $T$ when
selecting a minimum hitting set, this exponentially large number of
oracle calls is not observed.
(These differences between AbHS/AbHS+ and Hyper are experimentally
validated in~\autoref{sec:res}.)

%
%
It should be clear that the exponentially large reduction in the
number of oracle calls obtained with Hyper are hidden in the minimum
hitting set extractor. However, in Hyper the minimum hitting set
extractor is based on MaxSAT (concretely core-guided MaxSAT), and so
this hidden complexity is handled (most often efficiently) by the SAT
solver.

The inclusion of $M$ to find each set $S$ can also potentially save
exponentially many iterations.
%
%
Consider the following PAP \papdefn{2}:
\begin{equation} \label{eq:expex2}
\begin{array}{ll}
  V = & \{ m, t_1, x_1, \ldots, t_n, x_n \} \\
  H = & \{ (m\lor\neg x_1), (m\lor x_1\lor t_1), \\
  & (m\lor\neg x_n), (m\lor x_n\lor t_n) \} \\
  M = & \{ (m) \} \\
  T = & \{ (\neg t_1\lor\ldots\lor\neg t_n) \} \\
\end{array}
\end{equation}
In contrast with the previous example, $P$ has a solution,
i.e.\ there exists a subset $S$ of $H$ (with $S=H$) such that $T\land
S\entails M$.
Until the solution is found, all computed models of
$T\land S\land (\neg M)$ will also falsify $T\land S\land M$.
Any of these models might be filtered out if any candidate set $S$ is
such that $T\land S\land M$ is consistent.
It should be noted that in this case there is no formal guarantee that
the number of SAT oracle calls must be exponential. This depends on
the solutions provided by the minimum hitting set algorithm used.
Essentially, taking $M$ into account when selecting $S$ guarantees
that the picked set $S$ will not be such that $T\land S\entails\neg M$.
As the results in the next section confirm, in practice AbHS/AbHS+
can generate exponentially many candidates $S$ for which $T\land
S\entails\neg M$.

\begin{figure*}[!t]
  \begin{subfigure}[b]{\textwidth}
    \centering
    \includegraphics[width=\textwidth]{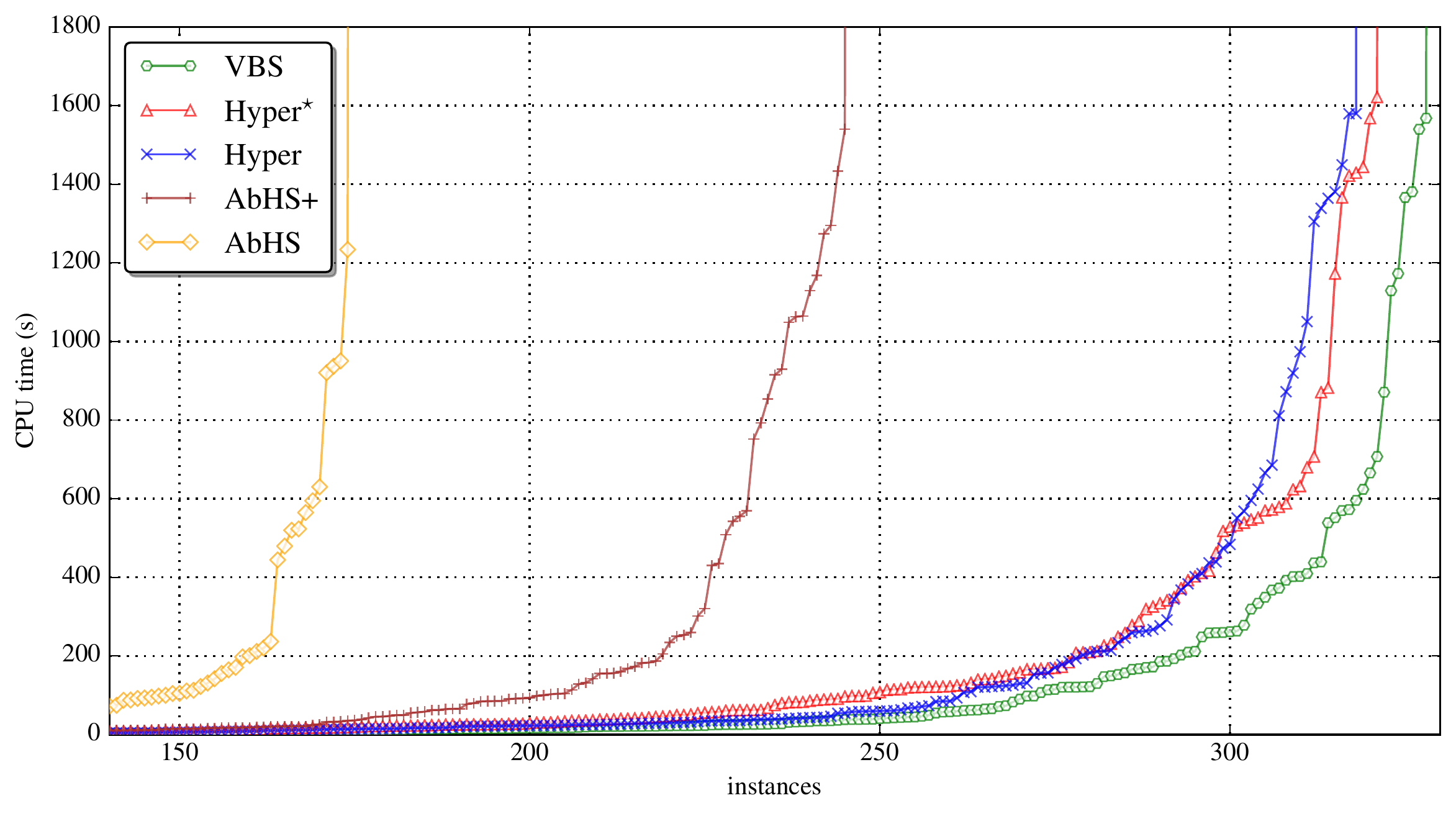}
    \caption{PMS instances}
    \label{fig:cactus-pms}
  \end{subfigure}%

  \vspace{1.0cm}

  \begin{subfigure}[b]{\textwidth}
    \centering
    \includegraphics[width=\textwidth]{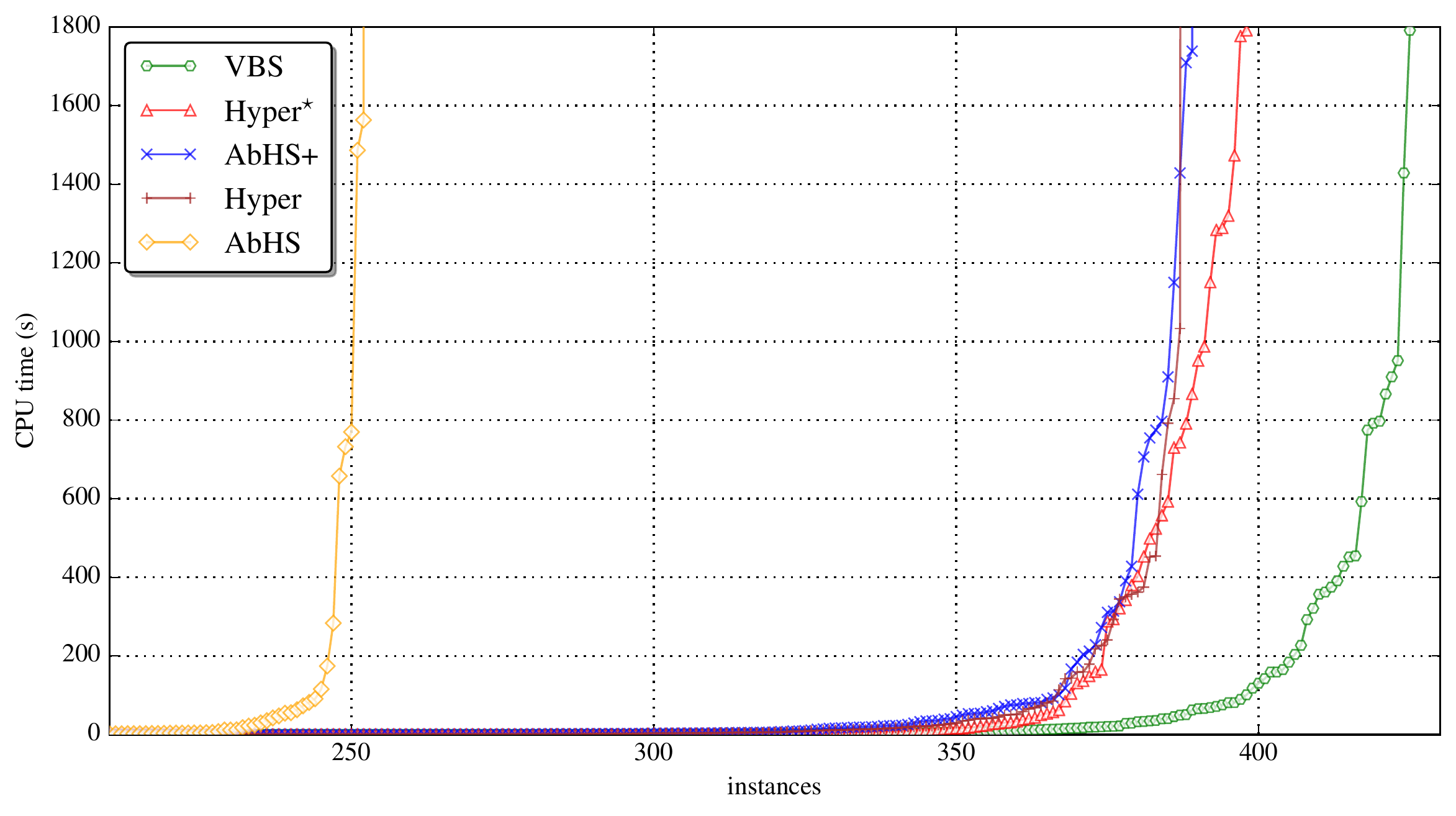}
    \caption{WPMS instances}
    \label{fig:cactus-wpms}
  \end{subfigure}
  \caption{Performance of Hyper, Hyper$^\star$, AbHS, and AbHS+ for the
  Abduction Problem Suite benchmarks.}
  \label{fig:cactus}
\end{figure*}

\section{Experimental Results} \label{sec:res}

This section evaluates the proposed approach to propositional abduction.

\subsection{Experimental Setup} \label{sec:res-setup}

All the conducted experiments were performed in Ubuntu Linux on an Intel
Xeon~E5-2630 2.60GHz processor with 64GByte of memory. The time limit was set
to 1800s and the memory limit to 10GByte for each process to run.
A prototype of the Hyper algorithm proposed above was implemented in C++ and
consists of two interacting parts. One of them computes minimum size hitting
sets of the set of counterexamples, also satisfying $T\land S\land M$. This is
achieved with the use of an incremental implementation of the algorithm based
on soft cardinality constraints~\cite{schaub-iclp12,mims-jsat15}, which is a
state-of-the-art MaxSAT algorithm that won several categories in the MaxSAT
Evaluation 2015\footnote{See results for MSCG15b at
\url{http://www.maxsat.udl.cat/15/}}.
The other part of the prototype checks satisfiability of $T\land
S\land(\neg{M})$, where $S$ is a candidate hitting set reported by the hitting
set solver.
Note that both parts of the solver were implemented on top of the well-known
SAT solver
Glucose~3.0\footnote{\url{http://www.labri.fr/perso/lsimon/glucose}}~\cite{audemard-sat13}.

Besides the basic version of Hyper, we also implemented an improved version,
which is below referred to as \hyperp and contains the first two improvements
described in \autoref{sec:opt}.
Namely, the first improvement does partial reduction of counterexamples by
traversing and trying to satisfy $0.2\times m$ clauses of each counterexample,
where $m$ is the size of the counterexample. The second improvement used in
\hyperp consists in bootstrapping the hitting set solver with 100 MCSes of
MaxSAT formula $T\land H\land (\neg{M})$. It should be noted that bootstrapping
the algorithm is not necessary but in some cases it can boost the performance
of the main algorithm.
Also note that the MaxSAT solver in both Hyper and \hyperp trims unsatisfiable
cores~\cite{mims-jsat15} detected during the solving process at most 5 times.

The performance of Hyper and \hyperp was compared to the performance of the
recent state-of-the-art algorithms AbHS and
AbHS+\footnote{\url{http://cs.helsinki.fi/group/coreo/abhs/}}\cite{jarvisalo-kr16}.
Additionally, we also implemented the QMaxHS approach described
in~\autoref{ssec:ihs}. The implementation was done on top of
DepQBF\footnote{\url{http://lonsing.github.io/depqbf/}}, the known QBF solver
which is capable or reporting unsatisfiable cores~\cite{egly-sat15}. However,
the performance of QMaxHS is poor (i.e.\ in our evaluation it did not solve
any instance from the considered benchmark suite) and we decided to exclude it
from consideration.

\subsection{Abduction Problem Suite} \label{sec:res-suite}
\begin{figure*}[!t]
  \begin{subfigure}[b]{0.49\textwidth}
    \centering
    \includegraphics[width=\textwidth]{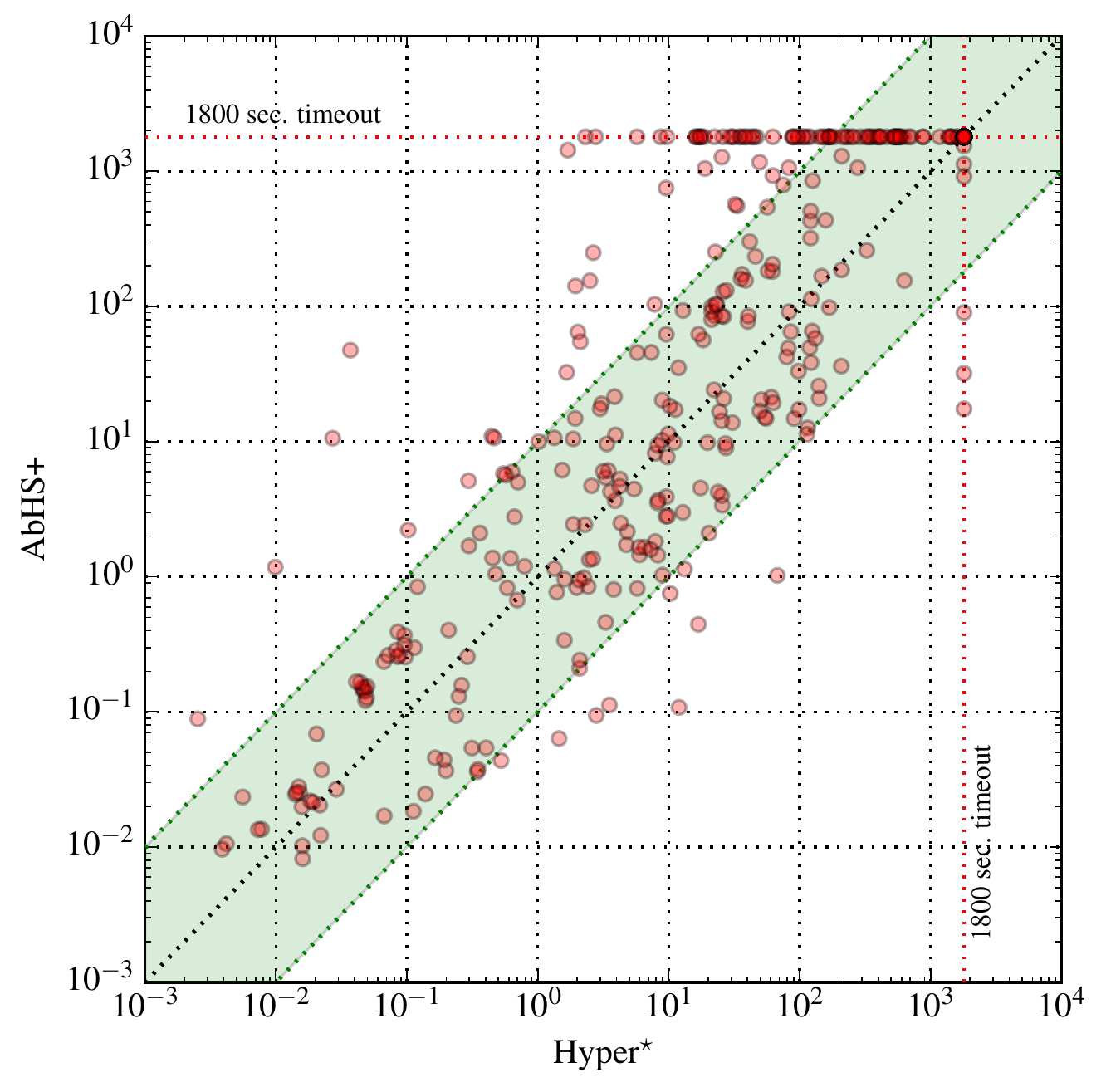}
    \caption{PMS instances}
    \label{fig:scatter-pms}
  \end{subfigure}%
  \begin{subfigure}[b]{0.49\textwidth}
    \centering
    \includegraphics[width=\textwidth]{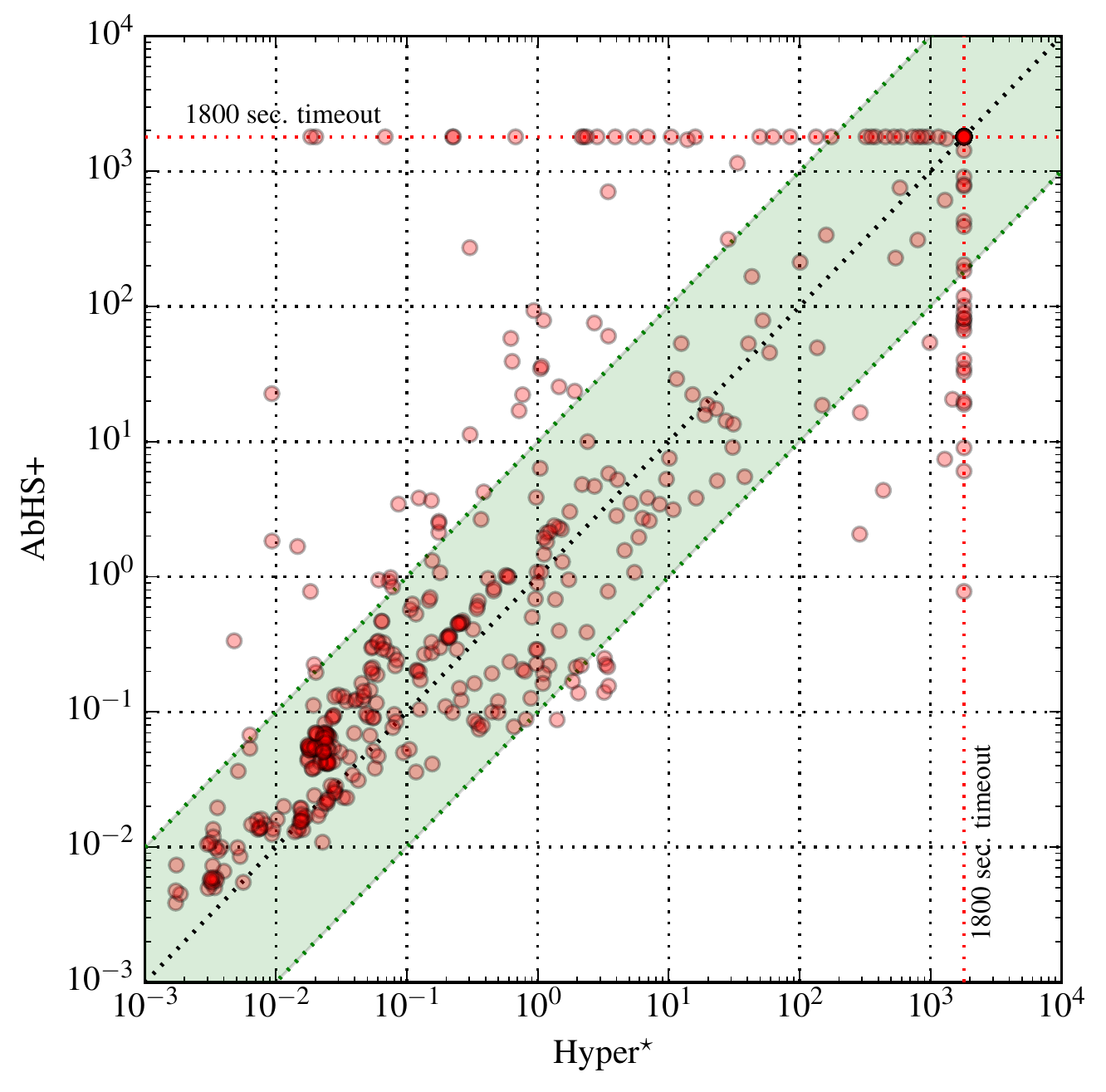}
    \caption{WPMS instances}
    \label{fig:scatter-wpms}
  \end{subfigure}
  \caption{\hyperp vs AbHS+.}
  \label{fig:scatter}
\end{figure*}

In order to assess the efficiency of the new approach to propositional
abduction, the following benchmark suite was used, which was proposed and also
considered in~\cite{jarvisalo-kr16}. According to~\cite{jarvisalo-kr16}, the
benchmark instances were generated based on crafted and industrial instances
from MaxSAT Evaluation 2014 with the use of the MaxSAT solver
LMHS\footnote{\url{http://www.cs.helsinki.fi/
group/coreo/lmhs/}}\cite{saikko-sat16} and the backbone solver
minibones\footnote{\url{http://sat.inesc-id.pt/~mikolas/sw/minibones/}}\cite{jlms-aicom15}.
The reader is referred to~\cite{jarvisalo-kr16} for details. The resulting
benchmark suite contains 6 benchmarks sets: \emph{pms-5}, \emph{pms-10},
\emph{pms-15}, \emph{wpms-5}, \emph{wpms-10}, and \emph{wpms-15}, where the
number indicates the number of manifestations. In the conducted experimental
evaluation, benchmark sets were aggregated based on their type (weighted or
unweighted) and resulted in two benchmark sets: PMS and WPMS, having 847 and
795 instances, respectively. The total number of instances is 1642.

The cactus plots reporting the performance of the considered algorithms
measured for the considered problem instances is shown in \autoref{fig:cactus}.
As one can see in~\autoref{fig:cactus-pms}, for PMS benchmark instances, Hyper
and \hyperp perform significantly better than both AbHS and AbHS+. More
precisely, \hyperp solves 321 instances (out of 847), which is 76 more
($>31\%$) than the number of instances solved by AbHS+ (245). The second best
competitor is Hyper, which solves 318 instances being 3 instances behind
Hyper$^\star$. Finally, the worst performance is shown by AbHS, which solves
174 instances.

\begin{table*}[!t]
  \caption{The number of iterations and running time per solver for example
  family \eqref{eq:expex1}. Additionally, for AbHS/AbHS+ the number of
  iterations of type 2 is also shown (in parentheses). Value $n$ varies from
  1 to 10.}
  \scriptsize
  \begin{center}
    \begin{tabular}{ccccccccccc}
      \toprule
      & \textbf{1} & \textbf{2} & \textbf{3} & \textbf{4} & \textbf{5} & \textbf{6} & \textbf{7} & \textbf{8} & \textbf{9} & \textbf{10} \\
      \midrule
      \multirow{2}{*}{\textbf{AbHS}} & 11 (7) & 59 (49) & 363 (343) & 2401 (829) & --- & --- & --- & --- & --- & --- \\
      \cmidrule{2-11}
      & \tiny{0.0s} & \tiny{0.1s} & \tiny{3.4s} & \tiny{166.2s} & \tiny{$>$1800s} & \tiny{$>$1800s} & \tiny{$>$1800s} & \tiny{$>$1800s} & \tiny{$>$1800s} & \tiny{$>$1800s} \\
      \midrule
      \multirow{2}{*}{\textbf{AbHS+}} & 6 (2) & 14 (4) & 28 (8) & 51 (16) & 125 (32) & 388 (64) & 978 (128) & 2242 (256) & --- & --- \\
      \cmidrule{2-11}
      & \tiny{0.0s} & \tiny{0.0s} & \tiny{0.0s} & \tiny{0.0s} & \tiny{0.3s} & \tiny{2.8s} & \tiny{24.3s} & \tiny{180.3s} & \tiny{$>$1800s} & \tiny{$>$1800s} \\
      \midrule
      \multirow{2}{*}{\textbf{Hyper}} & 6 & 14 & 17 & 19 & 27 & 32 & 32 & 35 & 39 & 48 \\
      \cmidrule{2-11}
      & \tiny{0.0s} & \tiny{0.0s} & \tiny{0.0s} & \tiny{0.0s} & \tiny{0.0s} & \tiny{0.0s} & \tiny{0.0s} & \tiny{0.0s} & \tiny{0.0s} & \tiny{0.0s} \\
      \bottomrule
    \end{tabular}
    \label{tab:expex1}
  \end{center}
\end{table*}

\begin{table*}[!t]
  \caption{The number of iterations and running time per solver for example
  family \eqref{eq:expex2}. Value $n$ varies from 1 to 10.}
  \small
  \begin{center}
    \begin{tabular}{ccccccccccc}
      \toprule
      & \textbf{1} & \textbf{2} & \textbf{3} & \textbf{4} & \textbf{5} & \textbf{6} & \textbf{7} & \textbf{8} & \textbf{9} & \textbf{10} \\
      \midrule
      \textbf{AbHS /} & 5 & 11 & 21 & 54 & 133 & 350 & 878 & 1995 & --- & --- \\
      \cmidrule{2-11}
      \textbf{AbHS+}& 0.0s & 0.1s & 0.1s & 0.2s & 0.3s & 2.0s & 15.5s & 168.8s & $>$1800s & $>$1800s \\
      \midrule
      \multirow{2}{*}{\textbf{Hyper}} & 6 & 16 & 17 & 14 & 20 & 29 & 18 & 27 & 30 & 37 \\
      \cmidrule{2-11}
      & 0.0s & 0.0s & 0.0s & 0.0s & 0.0s & 0.0s & 0.0s & 0.0s & 0.0s & 0.0s \\
      \bottomrule
    \end{tabular}
    \label{tab:expex2}
  \end{center}
\end{table*}

In contrast to the unweighted problem instances, the performance of the new
algorithm for weighted instances is penalized by the use of MaxSAT for
computing minimal hitting sets. The reason is that the MaxSAT solver used in
Hyper does not exploit any modern and widely used heuristics to efficiently
deal with weights, e.g.\ Boolean lexicographic optimization~\cite{jpms-amai11}
or stratification~\cite{ansotegui-aij13}.\footnote{This conjecture is also
suggested by the average numbers of iterations done by \hyperp and AbHS+ for
the WPMS benchmarks, which are 69 and 229, respectively. Since \hyperp does
significantly fewer iterations (on average) and solves around the same number
of instances as AbHS+ does, we assume that the calls to the MaxSAT oracle are
harder (on average).} This can explain a similar performance shown by Hyper and
\hyperp compared to AbHS+.  More precisely, \hyperp is able to solve 398
instances (out of 795).  AbHS+ comes second solving 389 instances while Hyper
is 2 instances behind AbHS+ (387 solved). The worst performance is shown again
by AbHS, which solves 252 instances.

Regarding the performance of the \emph{virtual best solver} (VBS), the data for
both sets of benchmarks can be seen in~\autoref{fig:cactus} and it is the
following. For PMS, the VBS aggregating Hyper, \hyperp as well as
AbHS+\footnote{AbHS is excluded from the VBS since it does not contribute to
its performance.} is able to solve 328 instances, which is 7 more instances
than what \hyperp can solve alone. In contrast, for the WPMS instances the
picture is different: the VBS solves 425 instances, which is 27 and 36 more
instances than what \hyperp and AbHS+ can solve separately. This indicates that
\hyperp and AbHS+ complement each other in this case, which suggests building a
portfolio of the solvers for weighted instances.
The performance comparison between AbHS+ and \hyperp is detailed in the scatter
plots shown in~\autoref{fig:scatter} and also confirms this conclusion.

Although the experimental results for the abduction problem suite show clear
performance gain of the proposed Hyper algorithm over the state-of-the-art in
propositional abduction (AbHS+), it is important to mention that the benchmark
suite was generated (see~\cite{jarvisalo-kr16}) from the MaxSAT instances by
filtering out those of them that are hard for MaxHS-like MaxSAT solvers, i.e.\
MaxHS~\cite{bacchus-cp11,bacchus-cp13} and LMHS~\cite{saikko-sat16}. This fact
suggests that applying similar ideas for generating problem instances by
targeting MaxSAT formulas that are easier for another family of MaxSAT
algorithms, e.g.\ the core-guided algorithms based on soft cardinality
constraints~\cite{mims-jsat15} (recall that the MaxSAT solver in Hyper is one
of them), would result in even better performance of Hyper. Moreover, the
experimental results for the weighted benchmarks emphasize the importance of
applying modern techniques (e.g.\ Boolean lexicographic optimization and
stratification) targeting specifically weighted instances. Having implemented
such improvements, one could expect Hyper to perform better for problem
instances with weights.

\subsection{Oracle Calls in AbHS+} \label{sec:res-calls}

This section studies the number of iterations for the considered approaches for
the families of examples described in \autoref{sec:analysis}. For both examples
\eqref{eq:expex1} and \eqref{eq:expex2} we generated 10 instances varying size
$n$ of the instance from 1 to 10 in order to show how the number of iterations
grows with the growth of size $n$ for each approach.\footnote{It should be
  noted that the pseudo-code in \autoref{alg:abhs} (taken
  from~\cite{jarvisalo-kr16}), as well as the actual source code, needs to be
  modified for AbHS+ to produce correct results when a PAP does not have a
  solution, as is the case with example \eqref{eq:expex1}. When blocking
  previously computed hitting sets, AbHS+ can generate clauses both with
  positive literals and clauses with negative literals and, for a PAP without
  solution, it will eventually compute an empty hitting set, denoting that
  there is no solution to the constraints added as sets to hit. As a result,
  the pseudo-code (and the source) needs to test for the case when the minimum
  hitting set returned is empty, in which case it must return `no solution'.
  This fix was added to AbHS+ to get the results presented in this section.}

Recall that example \eqref{eq:expex1} aims at showing the importance of adding
the theory clauses into the hitting set solver by saving an exponential number
of iterations related to candidates that are not consistent with the theory. In
AbHS/AbHS+ the consistency check is done through the second SAT call and
results in a counterexample blocking the candidate (AbHS+ also blocks its
supersets). The idea of proposing example family \eqref{eq:expex1} is, thus, to
show that the number of iterations of this type (let us call them iterations of
type 2 because they are related with the 2nd SAT call) in AbHS and AbHS+ can be
exponentially larger than the number of iterations done by Hyper\footnote{In
order test the number of iterations, all the optimizations related to \hyperp
were turned off and, thus, the basic version of Hyper was considered instead.}.
Indeed, \autoref{tab:expex1} confirms this conjecture indicating that the
number of iterations of type 2 in AbHS+ grows exponentially with the growth of
$n$, i.e.\ it is exactly $2^n$ (see the values in parentheses), while the
number of iterations done by Hyper is negligible. The situation gets even more
dramatic for AbHS. (As one can see, the total number of iterations performed by
AbHS and AbHS+ grows even faster.) The running time spent by AbHS and AbHS+
also grows significantly with the growth of $n$. As a result, AbHS and AbHS+
cannot solve any instances for $n>4$ and $n>8$, respectively, within 1800
seconds.  Observe that Hyper reports the result for each $n$ immediately.

Regarding example \eqref{eq:expex2}, it shows the importance of adding $M$ into
the hitting set solver. \autoref{tab:expex2} confirms that it can save an
exponential number of iterations. As one can observe, the number of iterations
grows exponentially with growing value $n$ for AbHS and AbHS+. Note that in
this case AbHS and AbHS+ behave similarly to each other, which is why
\autoref{tab:expex2} does not have a separate row for AbHS+. Analogously to the
previous case, the performance of the solver (i.e.\ its running time) is
severely affected by the number of iterations. Analogously to the previous
example and in contrast to AbHS and AbHS+, the basic version of Hyper does
significantly fewer iterations and spends almost no time for each of the
considered instances.


%
%
%

\section{Conclusions} \label{sec:conc}

Abduction finds many applications in Artificial Intelligence, with a
large body of work over the years. 
Recent work investigated propositional abduction, and proposed the use
of a variant of the implicit hitting set algorithm MaxHS for solving
the problem~\cite{jarvisalo-kr16}.

This paper identifies several sources of inefficiency with earlier
work, and proposes a novel, implicit hitting set inspired, algorithm
for propositional abduction. The novel algorithm, Hyper, is shown to
outperform the recently proposed algorithms AbHS and AbHS+ on existing
problem instances.
In addition, the paper demonstrates that the proposed improvements can
result in exponential savings on the number of SAT oracle calls, which
helps explain the observed performance improvements.
In a broader context, this paper contributes to the recent body of
work on implicit hitting set algorithms, and identifies algorithmic
optimizations that can be significant in other contexts.

A number of research directions can be envisioned. These include
improvements to the MaxSAT solver used for computing minimum hitting
sets, as this represents the main bottleneck of the Hyper
algorithm. Additional work will involve applying Hyper to a larger
range of problem instances.




\bibliographystyle{abbrv}
\bibliography{paper}

\end{document}